\documentclass[11pt,a4paper]{article}
\usepackage[hyperref]{acl2019}
\usepackage{latexsym}
\usepackage{todonotes}
\usepackage{algorithm,algpseudocode}
\usepackage{url}
\usepackage{booktabs}
\usepackage{siunitx}

\newcolumntype{B}{S[table-format=2.1,table-auto-round]} % for BLEU scores
\newcolumntype{N}{S[table-format=2.1,table-space-text-post=M]} % for numbers with prefix k, M, etc.
\newcolumntype{P}{S[table-format=3,table-space-text-post=\%]} % for percentages
\usepackage{multirow}
\usepackage{amsmath,mathtools}
\DeclareMathOperator*{\argmin}{arg\,min}
\usepackage{txfonts}
\usepackage{tikz}
\usepackage{paralist}

\aclfinalcopy % Uncomment this line for the final submission
\title{Efficiency through Auto-Sizing: \\[2pt]%Shrinking Parameters in the Transformer Network for the
Notre Dame NLP's Submission to the WNGT 2019 Efficiency Task}% the Transformer Network: \\[2pt]  Shrinking Parameters for the WNGT 2019 Efficiency Task}

\author{Kenton Murray \qquad Brian DuSell \qquad David Chiang \\
  Department of Computer Science and Engineering \\
  University of Notre Dame \\
  {\tt \{kmurray4,bdusell1,dchiang\}@nd.edu\ } }

\date{}

\begin{document}
\maketitle
\begin{abstract}
  This paper describes the Notre Dame Natural Language Processing Group's (NDNLP) submission to the WNGT 2019 shared task \citep{wngt2019}. We investigated the impact of auto-sizing \citep{murrayauto,murray19autosizing} to the Transformer network \cite{vaswani2017attention} with the goal of substantially reducing the number of parameters in the model. Our method was able to eliminate more than 25\% of the model's parameters while suffering a decrease of only 1.1 BLEU.
\end{abstract}

\section{Introduction}

The Transformer network \citep{vaswani2017attention} is a neural sequence-to-sequence model that has achieved state-of-the-art results in machine translation. However, Transformer models tend to be very large, typically consisting of hundreds of millions of parameters. As the number of parameters directly corresponds to secondary storage requirements and memory consumption during inference, using Transformer networks may be prohibitively expensive in scenarios with constrained resources. For the 2019 Workshop on Neural Generation of Text (WNGT) Efficiency shared task \citep{wngt2019}, the Notre Dame Natural Language Processing (NDNLP) group looked at a method of inducing sparsity in parameters called auto-sizing in order to reduce the number of parameters in the Transformer at the cost of a relatively minimal drop in performance.

Auto-sizing, first introduced by \citet{murrayauto}, uses group regularizers to encourage parameter sparsity. When applied over neurons, it can delete neurons in a network and shrink the total number of parameters. A nice advantage of auto-sizing is that it is independent of model architecture; although we apply it to the Transformer network in this task, it can easily be applied to any other neural architecture.

NDNLP's submission to the 2019 WNGT Efficiency shared task uses a standard, recommended baseline Transformer network. Following \citet{murray19autosizing}, we investigate the application of auto-sizing to various portions of the network. Differing from their work, the shared task used a significantly larger training dataset from WMT 2014 \citep{bojar2014findings}, as well as the goal of reducing model size even if it impacted translation performance. Our best system was able to prune over 25\% of the parameters, yet had a BLEU drop of only 1.1 points. This translates to over 25 million parameters pruned and saves almost 100 megabytes of disk space to store the model.

\section{Auto-sizing}

Auto-sizing is a method that encourages sparsity through use of a group regularizer. Whereas the most common applications of regularization will act over parameters individually, a group regularizer works over groupings of parameters. For instance, applying a sparsity inducing regularizer to a two-dimensional parameter tensor will encourage individual values to be driven to 0.0. A sparsity-inducing group regularizer will act over defined sub-structures, such as entire rows or columns, driving the entire groups to zero. Depending on model specifications, one row or column of a tensor in a neural network can correspond to one neuron in the model.

Following the discussion of \citet{murrayauto} and \citet{murray19autosizing}, auto-sizing works by training a neural network while using a regularizer to prune units from the network, minimizing:
\begin{equation*}
\mathcal{L} = -\sum_{\text{$f, e$ in data}} \log P(e \mid f; W) + \lambda R(\|W\|).
\end{equation*}
 $W$ are the parameters of the model and $R$ is a regularizer. %For simplicity, assume that the parameters form a single matrix $W$ of weights.
Here, as with the previous work, we experiment with two regularizers:
\begin{align*}
R(W) &= \sum_i \left(\sum_j W_{ij}^2\right)^{\frac12} && (\ell_{2,1}) \\
R(W) &= \sum_i \max_j |W_{ij}| && (\ell_{\infty,1})
\end{align*}
The optimization is done using proximal gradient descent \citep{parikh+boyd:2014}, which alternates between stochastic gradient descent steps and proximal steps:
\begin{align*}
W &\leftarrow W - \eta \nabla \log P(e \mid f; w) \\
W &\leftarrow \argmin_{W'} \left(\frac1{2\eta} \|W-W'\|^2 + R(W') \right)
\end{align*}

\section{Auto-sizing the Transformer}

\begin{figure}
    \centering
    \includegraphics[width=7cm]{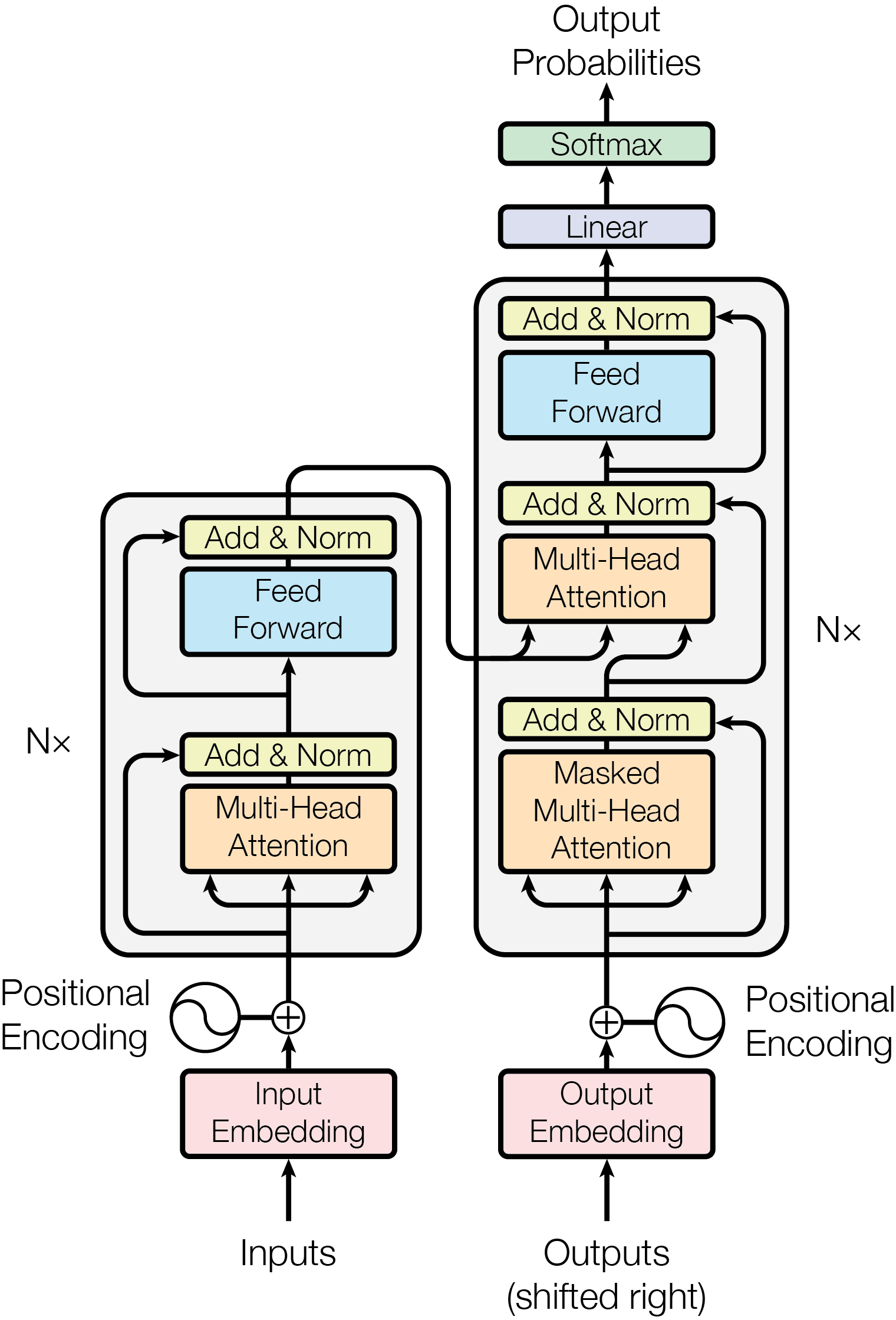}
    \caption{Architecture of the Transformer  \citep{vaswani2017attention}. We apply the auto-sizing method to the feed-forward (blue rectangles) and multi-head attention (orange rectangles) in all $N$ layers of the encoder and decoder. Note that there are residual connections that can allow information and gradients to bypass any layer we are auto-sizing. Following the robustness recommendations, we instead layer norm before.}
    \label{fig:transformer}
\end{figure}

The Transformer network \citep{vaswani2017attention} is a sequence-to-sequence model in which both the encoder and the decoder consist of stacked self-attention layers. The multi-head attention uses two affine transformations, followed by a softmax layer.
Each layer has a position-wise feed-forward neural network (FFN) with a hidden layer of rectified linear units.
Both the multi-head attention and the feed-forward neural network have residual connections that allow information to bypass those layers. In addition, there are also word and position embeddings. Figure \ref{fig:transformer}, taken from the original paper, shows the architecture. NDNLP's submission focuses on the $N$ stacked encoder and decoder layers.

The Transformer has demonstrated remarkable success on a variety of datasets, but it is highly over-parameterized. For example, the baseline Transformer model has more than 98 million parameters, but the English portion of the training data in this shared task has only 116 million tokens and 816 thousand types. Early NMT models such as \citet{sutskever2014sequence} have most of their parameters in the embedding layers, but the transformer has a larger percentage of the model in the actual encoder and decoder layers. Though the group regularizers of auto-sizing can be applied to any parameter matrix, here we focus on the parameter matrices within the encoder and decoder layers.

We note that there has been some work recently on shrinking networks through pruning. However, these differ from auto-sizing as they frequently require an arbitrary threshold and are not included during the training process. For instance, \citet{see2016compression} prunes networks based off a variety of thresholds and then retrains a model.
\citet{voita-etal-2019-analyzing} also look at pruning, but of attention heads specifically. They do this through a relaxation of an $\ell_0$ regularizer in order to make it differentiable. This allows them to not need to use a proximal step. This method too starts with pre-trained model and then continues training.
\citet{michel2019sixteen} also look at pruning attention heads in the transformer. However, they too use thresholding, but only apply it at test time. Auto-sizing does not require a thresholding value, nor does it require a pre-trained model.

Of particular interest are the large, position-wise feed-forward networks in each encoder and decoder layer:
\vspace{1mm}
\begin{equation*}
\text{FFN}(x) = W_2(\max(0,W_1x + b_1)) + b_2. 
\label{eq:ffn}
\end{equation*}
$W_1$ and $W_2$ are two large affine transformations that take inputs from $D$ dimensions to $4D$, then project them back to $D$ again. These layers make use of rectified linear unit activations, which were the focus of auto-sizing in the work of \citet{murrayauto}. No theory or intuition is given as to why this value of $4D$ should be used.

Following \cite{murray19autosizing}, we apply the auto-sizing method to the Transformer network, focusing on the two largest components, the feed-forward layers and the multi-head attentions (blue and orange rectangles in Figure \ref{fig:transformer}). Remember that since there are residual connections allowing information to bypass the layers we are auto-sizing, information can still flow through the network even if the regularizer drives all the neurons in a layer to zero -- effectively pruning out an entire layer.

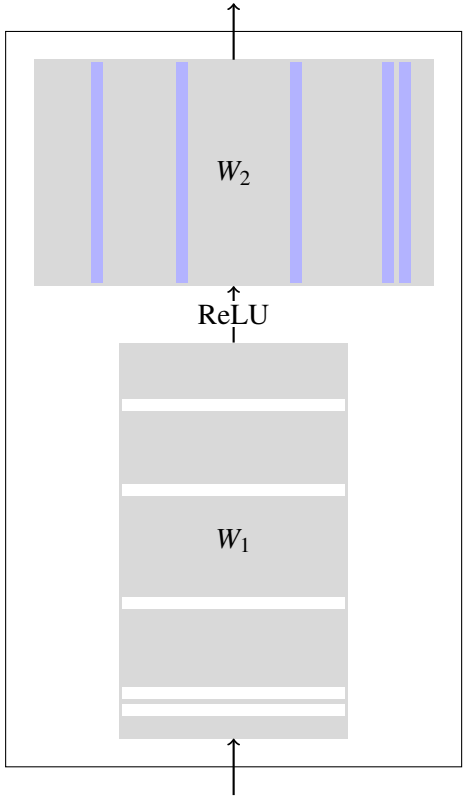
\begin{figure}
\tikzset{inner sep=1pt}
\begin{center}
\begin{tikzpicture}[scale=0.75]

\draw[fill=none] (-.5,-8.5) rectangle (7.5,4.5);

%Parameters
\draw[fill=gray!30,draw=none] (0,0) rectangle (7, 4);
\draw[fill=gray!30,draw=none] (1.5,-1) rectangle (5.5, -8);
\node[draw,draw=none] at (3.5,-4.5) {$W_1$};
\node[draw,draw=none] at (3.5,2) {$W_2$};
%\node[draw,text] at (3.5,-0.65) {ReLU};
%\draw[fill=gray!30,draw=none] (7.2,0) rectangle (7.4, 4);
%\draw[fill=gray!30,draw=none] (5.7,-1) rectangle (5.9, -8);

%Arrows
\draw[->,thick] (3.5,-1) to node[fill=white] {ReLU} (3.5,0);
\draw[->,thick] (3.5,-9) -- (3.5,-8);
\draw[->,thick] (3.5,4) -- (3.5,5);

%Zeroed Rows
\draw[fill=white!30,draw=none] (1.55,-2) rectangle (5.45, -2.2);
\draw[fill=white!30,draw=none] (1.55,-3.5) rectangle (5.45, -3.7);
\draw[fill=white!30,draw=none] (1.55,-5.5) rectangle (5.45, -5.7);
\draw[fill=white!30,draw=none] (1.55,-7.4) rectangle (5.45, -7.6);
\draw[fill=white!30,draw=none] (1.55,-7.1) rectangle (5.45, -7.3);

%To Delete Columns
\draw[fill=blue!30,draw=none] (1,0.05) rectangle (1.2, 3.95);
\draw[fill=blue!30,draw=none] (2.5,0.05) rectangle (2.7, 3.95);
\draw[fill=blue!30,draw=none] (4.5,0.05) rectangle (4.7, 3.95);
\draw[fill=blue!30,draw=none] (6.4,0.05) rectangle (6.6, 3.95);
\draw[fill=blue!30,draw=none] (6.1,0.05) rectangle (6.3, 3.95);

\end{tikzpicture}
\end{center}
\caption{Auto-sizing FFN network. For a row in the parameter matrix $W_1$ that has been driven completely to 0.0 (shown in white), the corresponding column in $W_2$ (shown in blue) no longer has any impact on the model. Both the column and the row can be deleted, thereby shrinking the model.}
\label{fig:ffn}
\end{figure}

\section{Experiments}

All of our models are trained using the fairseq implementation of the Transformer \cite{gehring2017convs2s}.\footnote{https://github.com/pytorch/fairseq}
%Following recommendations in both the fairseq and tensor-to-tensor code bases, we apply layer normalization before sub-components, which differs from the original paper.
For the regularizers used in auto-sizing, we make use of an open-source, proximal gradient toolkit implemented in PyTorch\footnote{https://github.com/KentonMurray/ProxGradPytorch} \citep{murray19autosizing}. For each mini-batch update, the stochastic gradient descent step is handled with a standard PyTorch forward-backward call. Then the proximal step is applied to parameter matrices.

%Hmmm... my numbers are different
%29.7 26.2 27.5 26.4 28.5 28.6 28.3 28.1

\begin{table*}
    \centering
    \begin{tabular}{r|c|c|c|c}
    \toprule
    System & Disk Size & Number of Parameters & newstest2014 & newstest2015 \\
    \hline
    Baseline & 375M & 98.2M & 25.3 & 27.9 \\
    \hline
    All $\ell_{2,1}=0.1$ & 345M & 90.2M & 21.6 & 24.1 \\
    \hline
    Encoder $\ell_{2,1}=0.1$ & 341M & 89.4M & 23.2 & 25.5 \\
    Encoder $\ell_{2,1}=1.0$ & 327M & 85.7M & 22.1 & 24.5 \\
    \hline
    FFN $\ell_{2,1}=0.1$ & 326M & 85.2M & 24.1 & 26.4 \\
    FFN $\ell_{2,1}=1.0$ & 279M & 73.1M & 24.0 & 26.8 \\
    FFN $\ell_{2,1}=10.0$ & 279M & 73.1M & 23.9 & 26.5 \\
    FFN $\ell_{\infty,1}=100.0$ & 327M & 73.1M & 23.8 & 26.0 \\
    \bottomrule
    \end{tabular}
    \caption{Comparison of BLEU scores and model sizes on newstest2014 and newstest2015. Applying auto-sizing to the feed-forward neural network sub-components of the transformer resulted in the most amount of pruning while still maintaining good BLEU scores.}
    \label{tab:scores}
\end{table*}

\subsection{Settings}

We used the originally proposed transformer architecture -- with six encoder and six decoder layers. Our model dimension was 512 and we used 8 attention heads. The feed-forward network sub-components were of size 2048. All of our systems were run using subword units (BPE) with 32,000 merge operations on concatenated source and target training data \cite{sennrich2016linguistic}. We clip norms at 0.1, use label smoothed cross-entropy with value 0.1, and an early stopping criterion when the learning rate is smaller than $10^{-5}$. We used the Adam optimizer \cite{kingma2014adam}, a learning rate of $10^{-4}$, and dropout of 0.1. Following recommendations in the fairseq and tensor2tensor \cite{tensor2tensor} code bases, we apply layer normalization before a sub-component as opposed to after. At test time, we decoded using a beam of 5 with length normalization \cite{boulanger2013audio} and evaluate using case-sensitive, tokenized BLEU \cite{papineni2002bleu}.

For the auto-sizing experiments, we looked at both $\ell_{2,1}$ and $\ell_{\infty,1}$ regularizers. We experimented over a range of regularizer coefficient strengths, $\lambda$, that control how large the proximal gradient step will be. Similar to \citet{murrayauto}, but differing from \citet{alvarez2016learning}, we use one value of $\lambda$ for all parameter matrices in the network. We note that different regularization coefficient values are suited for different types or regularizers. Additionally, all of our experiments use the same batch size, which is also related to $\lambda$. 

\subsection{Auto-sizing sub-components}

We applied auto-sizing to the sub-components of the encoder and decoder layers, without touching the word or positional embeddings. Recall from Figure \ref{fig:transformer}, that each layer has multi-head attention and feed-forward network sub-components. In turn, each multi-head attention sub-component is comprised of two parameter matrices. Similarly, each feed-forward network has two parameter matrices, $W_1$ and $W_2$. We looked at three main experimental configurations:

\begin{itemize}
    \item All: Auto-sizing is applied to every multi-head attention and feed-forward network sub-component in every layer of the encoder and decoder.
    \item Encoder: As with All, auto-sizing is applied to both multi-head attention and feed-forward network sub-components, but only in the encoder layers. The decoder remains the same.
    \item FFN: Auto-sizing applied only to the feed-forward network sub-components $W_1$ and $W_2$, but not to the multi-head portions. This too is applied to both the encoder and decoder.
\end{itemize}

%\subsubsection{Baseline}
%This architecture has 6 layers in both the encoder and decoder, each with 4 attention heads. Our model dimension is $d_{model}=512$, and our FFN dimension is 1024. 

\subsection{Results}

Our results are presented in Table \ref{tab:scores}. The baseline system has 98.2 million parameters and a BLEU score of 27.9 on newstest2015. It takes up 375 megabytes on disk. Our systems that applied auto-sizing only to the feed-forward network sub-components of the transformer network maintained the best BLEU scores while also pruning out the most parameters of the model. Overall, our best system used $\ell_{2,1}=1.0$ regularization for auto-sizing and left 73.1 million parameters remaining. On disk, the model takes 279 megabytes to store -- roughly 100 megabytes less than the baseline. The performance drop compared to the baseline is 1.1 BLEU points, but the model is over 25\% smaller.

Applying auto-sizing to the multi-head attention and feed-forward network sub-components of \emph{only} the encoder also pruned a substantial amount of parameters. Though this too resulted in a smaller model on disk, the BLEU scores were worse than auto-sizing just the feed-forward sub-components. Auto-sizing the multi-head attention and feed-forward network sub-components of both the encoder \emph{and} decoder actually resulted in a larger model than the encoder only, but with a lower BLEU score. Overall, our results suggest that the attention portion of the transformer network is more important for model performance than the feed-forward networks in each layer.

\section{Conclusion}

In this paper, we have investigated the impact of using auto-sizing on the transformer network of the 2019 WNGT efficiency task. We were able to delete more than 25\% of the parameters in the model while only suffering a modest BLEU drop. In particular, focusing on the parameter matrices of the feed-forward networks in every layer of the encoder and decoder yielded the smallest models that still performed well.

A nice aspect of our proposed method is that the proximal gradient step of auto-sizing can be applied to a wide variety of parameter matrices. Whereas for the transformer, the largest impact was on feed-forward networks within a layer, should a new architecture emerge in the future, auto-sizing can be easily adapted to the trainable parameters.

Overall, NDNLP's submission has shown that auto-sizing is a flexible framework for pruning parameters in a large NMT system. With an aggressive regularization scheme, large portions of the model can be deleted with only a modest impact on BLEU scores. This in turn yields a much smaller model on disk and at run-time.

\section*{Acknowledgements}
This research was supported in part by University of Southern California, subcontract 67108176 under DARPA contract HR0011-15-C-0115.

%\begin{itemize}
%\item Left and right margins: 2.5 cm
%\item Top margin: 2.5 cm
%\item Bottom margin: 2.5 cm
%\item Column width: 7.7 cm
%\item Column height: 24.7 cm
%\item Gap between columns: 0.6 cm
%\end{itemize}

\bibliography{acl2019}
\bibliographystyle{acl_natbib}

\end{document}